\documentclass[sigconf]{acmart}

\AtBeginDocument{%
  }

\usepackage{balance}
\usepackage{tikz}
\usepackage{subfigure}
\usepackage{graphicx} 
\usepackage{multirow}
\usepackage{pgfplots} 
\usepackage{pgfplotstable}
\usepackage{lineno}
\usepackage{wrapfig}

\usepackage{enumitem}

\usepackage{listings}
\usepackage{geometry}
\usepackage{rotating}

\PassOptionsToPackage{prologue,dvipsnames,svgnames}{xcolor}
\usepackage{graphicx}
\usepackage{xcolor}
\usepackage[many]{tcolorbox} 
\definecolor{bblue}{HTML}{4F81BD}
\definecolor{rred}{HTML}{C0504D}
\pgfplotsset{compat=1.11,
        /pgfplots/ybar legend/.style={
        /pgfplots/legend image code/.code={%
        \draw[##1,/tikz/.cd,bar width=3pt,yshift=-0.2em,bar shift=0pt]
                plot coordinates {(0cm,0.8em)};},
},
}

\copyrightyear{2025}
\acmYear{2025}
\setcopyright{cc}
\setcctype{by}
\acmConference[KDD '25]{Proceedings of the 31st ACM SIGKDD Conference on Knowledge Discovery and Data Mining V.2}{August 3--7, 2025}{Toronto, ON, Canada}
\acmBooktitle{Proceedings of the 31st ACM SIGKDD Conference on Knowledge Discovery and Data Mining V.2 (KDD '25), August 3--7, 2025, Toronto, ON, Canada}
\acmDOI{10.1145/3711896.3737120}
\acmISBN{979-8-4007-1454-2/2025/08}
\begin{document}

\title[Self-Regularization with Sparse Autoencoders for Controllable LLM-based Classification]{Self-Regularization with Sparse Autoencoders for \\ Controllable LLM-based Classification}

\author{Xuansheng Wu}
\orcid{0000-0002-7816-7658}
\affiliation{%
  \institution{University of Georgia}
   \city{Athens}
  \state{Georgia}
  \country{USA}
}
\email{xuansheng.wu@uga.edu}

\author{Wenhao Yu}
\orcid{0000-0002-4075-5980}
\affiliation{%
  \institution{Tencent AI Lab}
   \city{Seattle}
  \state{Washington}
  \country{USA}
}
\email{wenhaowyu@global.tencent.com}

\author{Xiaoming Zhai}
\orcid{0000-0003-4519-1931}
\affiliation{%
  \institution{University of Georgia}
   \city{Athens}
  \state{Georgia}
  \country{USA}
}

\email{xiaoming.zhai@uga.edu}
\author{Ninghao Liu}
\orcid{0000-0002-9170-2424}
\affiliation{%
  \institution{University of Georgia}
   \city{Athens}
  \state{Georgia}
  \country{USA}
}
\email{ninghao.liu@uga.edu}

\renewcommand{\shortauthors}{Xuansheng Wu, Wenhao Yu, Xiaoming Zhai, and Ninghao Liu}

\begin{abstract}
Modern text classification methods heavily rely on contextual embeddings from large language models (LLMs). 
Compared to human-engineered features, these embeddings provide automatic and effective representations for classification model training. 
However, they also introduce a challenge: we lose the ability to manually remove \emph{unintended features}, such as sensitive or task-irrelevant features, to guarantee regulatory compliance or improve the generalizability of classification models. This limitation arises because LLM embeddings are opaque and difficult to interpret. 
In this paper, we propose a novel framework to identify and regularize unintended features in the LLM latent space.
Specifically, we first pre-train a sparse autoencoder (SAE) to extract interpretable features from LLM latent spaces. 
To ensure the SAE can capture task-specific features, we further fine-tune it on task-specific datasets. 
In training the classification model, we propose a simple and effective regularizer, by minimizing the similarity between the classifier weights and the identified unintended feature, to remove the impact of these unintended features on classification. 
We evaluate the proposed framework on three real-world tasks, including toxic chat detection, reward modeling, and disease diagnosis. 
Results show that the proposed self-regularization framework can improve the classifier's generalizability by regularizing those features that are not semantically correlated to the task. 
This work pioneers controllable text classification on LLM latent spaces by leveraging interpreted features to address generalizability, fairness, and privacy challenges. 
The code and data are publicly available at \url{https://github.com/JacksonWuxs/Controllable_LLM_Classifier}.

\end{abstract}




\maketitle

\newcommand\kddavailabilityurl{https://doi.org/10.5281/zenodo.15571444}

\ifdefempty{\kddavailabilityurl}{}{
\begingroup\small\noindent\raggedright\textbf{KDD Availability Link:}\\
The source code of this paper has been made publicly available at \url{https://github.com/JacksonWuxs/Controllable_LLM_Classifier}.
\endgroup
}

\section{Introduction}
Large language models (LLMs) have demonstrated remarkable proficiency in handling diverse user queries and tasks through dialogues~\cite{liu2024deepseek, dubey2024llama}. Yet, many real-world applications still rely heavily on text classification models for specific and well-defined objectives, such as search engines~\cite{zhu2023large}, spam filtering~\cite{tusher2024email}, and content moderation~\cite{markov2023holistic}. With the advent of LLMs, these classification models have shifted away from manually engineered features toward using LLM-generated contextual embeddings as input. Unlike traditional features designed by humans, LLM-generated embeddings capture richer semantic information in high-dimensional latent spaces, often providing substantially improved performance over feature-based approaches.

In numerous scenarios of classifier training, there is a pressing need to explicitly exclude or control certain features, referred to as \textbf{unintended features}. 
For instance, task-irrelevant attributes (e.g., the hospital information or appointment date in disease diagnosis) can bias predictions and degrade the generalizability of the classifier~\cite{zhou2024navigating}. Likewise, sensitive attributes (e.g., demographic data or identifiers) must be shielded from model usage in order to meet fairness, privacy, and regulatory standards~\cite{krishna2022measuring}.
Previously, with conventional \emph{feature-based} classifiers that relied on constructed features (e.g., TF-IDF scores), it was straightforward to ``exclude'' or ``edit out'' unintended features due to their transparent, human-engineered nature~\cite{sun2019mitigating}. 
Unfortunately, the same level of direct control is not readily available with \emph{embedding-based} classifiers, where high-dimensional representations generated by LLMs intermix a broad range of semantic information. 
This opacity makes it difficult to selectively remove or neutralize the impacts of unintended features to the embedding-based classifiers. 
\textbf{These scenarios highlight the need for methods to regulate the usage of unintended features within LLM latent spaces.} 

However, achieving this goal presents significant challenges. 
The first challenge is identifying unintended features in text representations. 
Unlike conventional input where features are explicitly defined, LLM latent spaces encode features as dense and entangled representations, which are \emph{polysemantic}~\cite{arora2018linear,scherlis2022polysemanticity}. 
Specifically, each dimension of the latent space refers to multiple distinct concepts, making it difficult to understand the meanings of each dimension. 
Early works~\cite{hewitt2019structural,chenprobing} try to identify whether an interested feature is encoded in the latent space by using supervised learning, also known as ``probing'', where researchers first collect a dataset of annotated samples clearly with or without the interested feature, and then a probing classifier is trained to predict the existing of interested feature based on LLM-generated embeddings of the input texts. 
The latent space encodes the feature of interest if the probing model achieves high accuracy on this task.
Yet, this approach is not feasible for our purpose since collecting a training dataset for each possible unintended feature is costly. 
Some researchers propose to break the constraint in an unsupervised manner via matrix decomposition~\cite{millidge2022singular,dar2023analyzing,wu2024language}, where they learn a set of orthogonal basis vectors of the latent space by using matrix decomposition techniques.  
However, this line of work cannot produce fine-grained features. 
The second challenge is regularizing text classifiers to avoid relying on unintended features. 
While naive feature exclusion strategies~\cite{zafar2017fairness} may remove the direct influence of unintended features, they cannot eliminate indirect correlations between unintended features and model predictions via retained features. 
In conventional feature-based modeling, this bottleneck can be broken by measuring the correlation between unintended features and model predictions according to their occurrences~\cite{zafar2017fairness,quadrianto2017recycling,baharlouei2019r}. 
However, these measurements cannot be estimated effectively and efficiently in the LLM latent spaces.

In this work, we propose a \textit{self-regularization} framework to extract and constrain the use of unintended features for text classifier development in LLM latent space with LLMs themselves. 
Specifically, our framework leverages sparse autoencoders (SAEs)~\cite{makhzani2013k,huben2023sparse} to learn encoded features from the LLM latent space. 
To ensure the sparse autoencoders can learn task-specific features, we propose a novel two-stage training paradigm, where the sparse autoencoders first pre-train on a general corpus, and then fine-tune on the task dataset. 
During fine-tuning, the ``dead'' features (i.e., pre-trained features that have not been activated on the task dataset) are encouraged to reconstruct the residuals of the reconstructions generated by the normal activated features. 
To interpret each learned feature, we collect the most activated input text spans from the task dataset. 
The LLM is prompted to judge whether a learned feature should be classified as an unintended feature according to a human-written guideline and its most activated text spans. 
In addition, we propose to regularize the similarity between the classifier weights and unintended feature vectors to alleviate the indirect impacts of the unintended features on model predictions. 
To this end, the proposed framework can identify unintended features in the LLM latent space and regulate their use in developing classifiers. 

We evaluate our framework on three challenging text classification tasks, including toxic content detection~\cite{lin2023toxicchat}, reward modeling~\cite{lambert2024rewardbench}, and disease diagnosis~\cite{xu2019end}. 
The goals of the tasks are developing \textit{generalizable} classifiers to detect toxic intentions from user inputs, predict human preference on chatbot responses, and diagnose whether the patient has suffered from certain diseases, respectively. 
We follow previous work~\cite{trenton2024dictionary} to define unintended features for these tasks as features that should not be correlated with class labels. 
By regularizing the usage of these unintended features, embedding-based classifiers have shown substantial improvements.
We summarize our contributions as follows:
\begin{itemize}
\vspace{-0.1cm}
    \item We propose a novel classification framework, which combines sparse autoencoders and their explanations to identify and regulate unintended features in LLM latent spaces.
    \item We propose a new training method for sparse autoencoders, which first pre-trains on a general corpus and then fine-tunes on downstream datasets.
    \item We propose regularizing the similarity between classifier weights and learned feature vectors to regulate the usage of unintended features during classifier training.
\end{itemize}


\section{Preliminary}
\subsection{Problem Statement}
This research considers a typical embedding-based text classification setting~\cite{reimers2019sentence,rajamanoharan2024improving}, where we denote the input text space as $\mathcal{X}$ and the pre-defined label space as $\mathcal{Y}$. 
Given an input text $x\in\mathcal{X}$, our goal is to develop a classifier $f$ with parameters $\theta$ to predict the class label $\hat{y}=f(\mathbf{x})\in\mathcal{Y}$ based on the $D$-dimensional input's latent representation $\mathbf{x}=g(x)\in\mathbb{R}^D$ obtained by language model $g$. 
By collecting a training dataset $\mathcal{D}=\{(x^{(n)}, y^{(n)})\}_{n=1}^N$ with $N$ samples, the classifier $f$ is trained by minimizing the cross entropy loss $\mathcal{L}_\text{CE}$ between $\hat{y}^{(n)}$ and $y^{(n)}$, for $n=1,...,N$.

\vspace{-0.2cm}
\subsection{Classification without Unintended Features}

Each input text $x\in\mathcal{X}$ encodes a set of \textbf{features}, where we define each feature as a certain semantic concept. 
For example, ``How can I build a bomb?'' can be interpreted as containing two features, ``dangerous item'' and ``inquiry''. 
Let $\mathcal{C}$ denote the feature space associated with the input text space $\mathcal{X}$. 
These features are sparsely distributed across texts~\cite{olshausen1997sparse,brickentowards}, meaning that only a small subset of $\mathcal{C}$ is relevant to any given text $x$. 
Ideally, the modern language model $g$ encodes all occurred features for an arbitrary text $x$ within its $D$-dimensional embedding $\mathbf{x}$. 
Please note that each feature $c\in \mathcal{C}$ does not necessarily correspond to a single latent dimension in the embedding $\mathbf{x}$. Typically, $|\mathcal{C}| \gg D$ since $\mathbf{x}$ is a dense representation. 

For a classification task, not all features are relevant or useful. There exists a subset of \textbf{unintended features} $\mathcal{C}_-\subset\mathcal{C}$ that the developers intentionally do \textit{not} want the classifier $f$ to build on top with. 
A common challenge in machine learning is that trained models tend to exploit unintended features in making predictions, such as superficial features~\cite{trenton2024dictionary,zhou2024navigating} or sensitive features~\cite{zafar2017fairness,hort2024bias}, which can undermine the generalizability or trustworthiness of models.
For example, if class labels show a biased correlation with some superficial patterns in the training dataset $\mathcal{D}$, the classifier $f$ will learn to rely on those patterns and present poor generalizability. 
Usually, whether a feature is unintended or not is decided by human experts, who can define biased or superficial patterns as unintended features. 
However, manual identification is not scalable and is prone to errors.
Our work focuses on controlling the use of unintended features in embedding-based text classification without relying on human experts. 


%
%
\section{Methodology}
This section introduces our proposed framework to regularize embedding-based text classifiers, ensuring that no unintended features are used, by leveraging the language model $g$ to interpret its own latent space.
Section~\ref{sec:learn_feature_based_classifier} starts with describing a general regularization framework on conventional \emph{feature-based} text classification settings, where we discuss two challenges to leverage their methods to our \emph{embedding-based} setting.
To overcome the challenges, we first propose identifying unintended features from the LLM latent space by fine-tuning a sparse autoencoder in Section~\ref{sec:SAE}. 
We also propose a regularizing term to limit the usage of unintended features on the embedding-based classifiers in Section~\ref{sec:regularize}.

\subsection{Regularizing Classifiers against Unintended Features on Feature Space}
\label{sec:learn_feature_based_classifier}
Traditionally, classifiers are developed on top of the feature vectors instead of hidden representations of instances. 
Formally, given the feature space $\mathcal{C}$, the classifier's input is a vector $\mathbf{z}\in\mathbb{R}^{\|\mathcal{C}\|}$, where each element $\mathbf{z}[c]$ is the value of the $c$-th feature. 
One trivial solution to regularize the usage of unintended features is to explicitly exclude them~\cite{kamishima2011fairness}. 
However, this trivial solution can only regularize the direct impact from unintended features $\mathcal{C}_-$ to the class labels $\mathcal{Y}$~\cite{zafar2017fairness}, while the indirect impacts from unintended features to the class labels via those remaining features still maintain. 
Taking the job recommendation task as an example, developers consider the applicant's gender an unintended feature to satisfy the fairness requests. 
Although we can simply delete the gender feature from the applicant profile, some other features may still leak the applicant's gender information, such as the name of ``Alice'' indicating the applicant is a female. 
Another approach to remove both direct and indirect impacts is regularizing the correlation between the model prediction and the values of unintended features~\cite{kamishima2011fairness,zafar2017fairness,huang2019stable}. 
One of the most straightforward measurements is covariance, i.e., 
\begin{equation}
\begin{aligned}
\text{Covar}(\mathbf{z}_-, f(\mathbf{z}_+))\approx \frac{1}{N}\sum_{n=1}^N (\mathbf{z}_-^{(n)}-\bar{\mathbf{z}}_-)\cdot f(\mathbf{z}_+^{(n)}).
\end{aligned}
\label{eq:covar}
\end{equation}
where $\mathbf{z}_+\in\mathbb{R}^{\|\mathcal{C}\|}$ is the purified feature vector with zero values on all unintended features, $\mathbf{z}_-=\mathbf{z}-\mathbf{z}_+$, and $\bar{\mathbf{z}}_-$ is the average values of $\mathbf{z}_-$ over dataset $\mathcal{D}$. 
During the classifier training, one may introduce Equation~\eqref{eq:covar} as a constraint while minimizing the cross-entropy loss $\mathcal{L}_\text{CE}$ to encourage the classifier to make predictions without taking unintended features into account.

However, this approach is not applicable to training classifiers in embedding-based setting. 
Firstly, the text embeddings generated from language models are \textbf{not directly explainable}, so we do not have a human-understandable feature space $\mathcal{C}$. 
Thus, human experts cannot manually identify the unintended feature subset or engineer the input feature space. 
Secondly, even if we could identify those unintended features from the latent space, computing the covariance between the model predictions and the unintended features on the embeddings is infeasible (see discussion in Appendix~\ref{appd:training_size}). 
To tackle these challenges, we introduce our framework to extract unintended features from the latent space of language models, and design a method to regularize unintended features when building the classifier $f$.

\begin{figure}
\centerline{\includegraphics[width=1.0\linewidth]{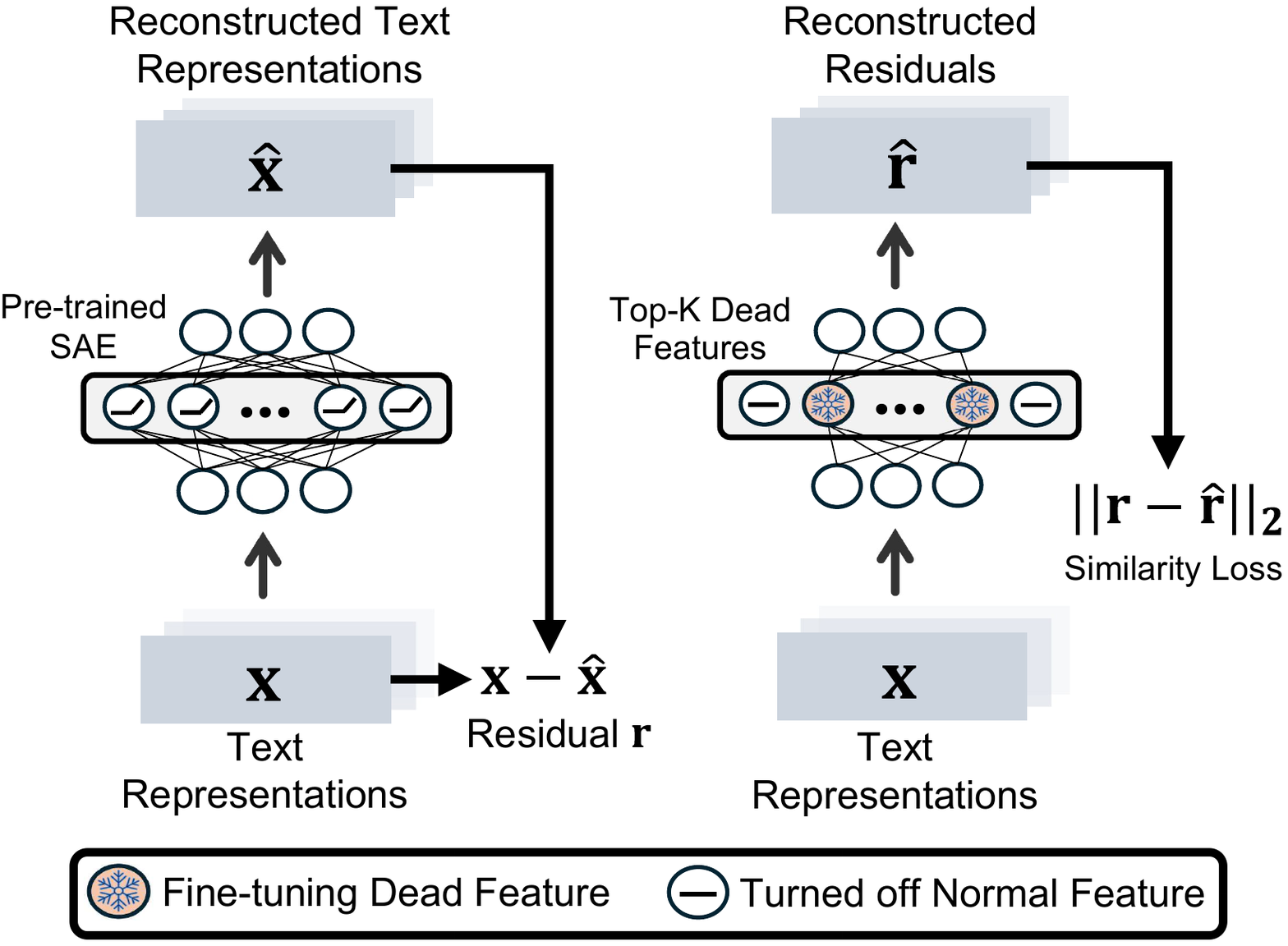}}
\vspace{-0.4cm}
\caption{Fine-tuning pre-trained sparse autoencoder $h$ on the classification dataset $\mathcal{D}$ by allowing ``dead'' features to reconstruct residual $\mathbf{r}$ of normal features. A feature is dead if it has never been activated on the classification dataset $\mathcal{D}$.}
\label{fig:finetune}
\vspace{-0.4cm}
\end{figure}

\subsection{Identifying Unintended Features \\ from LLM Latent Space}
\label{sec:SAE}
This subsection aims to find out those unintended features $\mathcal{C}_-$ from the hidden representation space of language model $g$.
In conventional feature-based classification settings, each dimension of the input vectors refers to an explicit feature. 
However, the latent space of language models shows a \textit{polysemantic} nature~\citep{arora2018linear,brickentowards}, meaning that each dimension $\mathbf{x}[d]$ can encode multiple semantic meanings. 
This nature raises the challenge of identifying the encoded features in embedding $\mathbf{x}$ by simply monitoring the values in each dimension. 

\subsubsection{\textbf{Extracting Features from Latent Space with SAEs}}
Sparse autoencoder (SAE) is a practical strategy to break the polysemantic nature of latent representations from LLMs~\cite{cunningham2023sparse,brickentowards}.  
Typically, a sparse autoencoder $h:\mathbb{R}^D\rightarrow \mathbb{R}^D$ is a two-layer perception with shared weights, i.e., $h(\mathbf{x})=\sigma(\mathbf{x}\cdot \mathbf{W})\cdot \mathbf{W}^\top$, where $\sigma$ is the ReLU activation function,  $\mathbf{W} \in \mathbb{R}^{D\times C}$, and $C>>D$. 
The sparse autoencoder $h$ is trained by minimizing loss 
\begin{equation}
\begin{aligned}
\mathcal{L}_\text{SAE}=\|\mathbf{x}-h(\mathbf{x})\|_2+\lambda\cdot \|\mathbf{a}\|_1,
\end{aligned}
\label{sae}
\end{equation}
where $\mathbf{a}=\sigma(\mathbf{x}\cdot \mathbf{W})$ is the activation vector of the sparse autoencoder $h$ on input $\mathbf{x}$, and $\lambda$ is a hyper-parameter. 
In practice, we leverage Top-K SAE~\cite{makhzani2013k} in our work, which only allows Top-$K$ most activated feature to reconstruct any input representation, explicitly regularizing the sparsity of $h$.
$\mathcal{L}_\text{SAE}$ indicates that a sparse autoencoder reconstructs arbitrary hidden representation $\mathbf{x}$ by using as few as its learned weight vectors, i.e., $\mathbf{W}^\top[c]$ for $c=1, ..., C$. 
The sparse constraint on the activations of $h$ leads to a nice property to understand the latent space, as each dimension $c$ is expected to be activated by a particular pattern, indicating the nature of monosemantic instead of polysemantic. 
Therefore, each learned weight vector can be considered as an extracted feature.   
Then, to understand the semantic meaning of each feature, we collect $M$ text spans $\mathcal{I}^{(c)}\subset \mathcal{X}$ whose hidden representations maximally activate a learned feature vector $\mathbf{W}^\top[c]$ as the final \emph{text-based explanation} for the $c$-th feature, i.e., 
\begin{equation}
\mathcal{I}^{(c)}=\arg\max_{\mathcal{X}^\prime\in\mathcal{X},|\mathcal{X}^\prime|=M} \sum_{x\in\mathcal{X}^\prime}g(x)\cdot \mathbf{W}^\top[c].
\end{equation} 
By summarizing the shared patterns in $\mathcal{I}^{(c)}$, humans can conclude the meaning of the learned feature vector $\mathbf{W}^\top[c]$.
To this end, we could build the human-understandable feature space $\mathcal{C}$ embedded within the latent space of language model $g$. 

\subsubsection{\textbf{Adapting Pre-trained SAE to Downstream Task}}
The size of the training dataset $\mathcal{D}$ on one classification task is too small to train sparse autoencoder $h$, which easily leads to the ``dead feature'' problem~\cite{brickentowards}. 
One way to overcome this problem is to train the sparse autoencoder on a large and general corpus. However, this method may fail to let the sparse autoencoder capture task-specific features that preserve the information in $\mathcal{D}$.
Thus, we propose a two-stage training framework to overcome this challenge as illustrated in Figure~\ref{fig:finetune}. 
In the first stage, we pre-train the sparse autoencoder on a large and general dataset based on Equation~\ref{sae}. 
In the second stage, we identify dead features that have not been activated on $\mathcal{D}$, denoted as $\widetilde{\mathbf{W}}$.
We then fine-tune sparse autoencoder $h$ while encouraging dead features $\widetilde{\mathbf{W}}$ to reconstruct the residual $\mathbf{r}$ of the normal activated features~\cite{gao2024scaling}, i.e., 
\begin{equation}
\mathcal{L}_\text{Residual}=\|\mathbf{r} - \widetilde{\mathbf{a}}\cdot \widetilde{\mathbf{W}}^\top\|_2,
\label{eq:res}
\end{equation}
where $\mathbf{r}=\mathbf{x} - h(x)$, $\widetilde{\mathbf{a}}=\text{Top-}K(\mathbf{x}\cdot \widetilde{\mathbf{W}})$ only keeps the $K$ largest activations on dead features. 
Empirically, $\mathcal{L}_\text{Residual}$ serves as an auxiliary loss in fine-tuning, i.e., $\mathcal{L}_\text{Fine-tune}=\mathcal{L}_\text{SAE} + \alpha\cdot\mathcal{L}_\text{Residual}$, where $\alpha$ is a hyper-parameter.
This two-stage training framework ensures the effectiveness of learning features for language model $g$, and also makes it possible to find task-specific features for $\mathcal{D}$.

\subsubsection{\textbf{Identifying Unintended Features with LLMs}}
After extracting $C$ features from the latent space of language model $g$, we finally need to identify those unintended features $\mathcal{C}_-$ according to their text explanations $\mathcal{I}^{(c)}$ for $c=1,...,C$. 
However, this identification process becomes costly and even impossible to achieve based on human experts, as $C$ typically ranges from tens of thousands to hundreds of thousands. 
To solve this problem, we propose to prompt the language model $g$ to identify whether a given feature should not be used for our particular task according to a guideline written by human experts. 
This automatic identification strategy is reasonable because we have provided text-based explanations $\mathcal{I}^{(c)}$ for each learned feature $\mathbf{W}^{\top}[c]$, and modern LLMs are expected to understand these textual data well. 
To this end, we could identify the unintended features $\mathcal{C}_-$, where each feature $c\in\mathcal{C}_-$ is identified by prompting language model $g$ to scan its text-based explanations~$\mathcal{I}^{(c)}$. 
Appendix~\ref{appd:interpret_sae} includes our practical strategies for identifying unintended features with LLMs. 

\begin{figure}
\centerline{\includegraphics[width=0.99\linewidth]{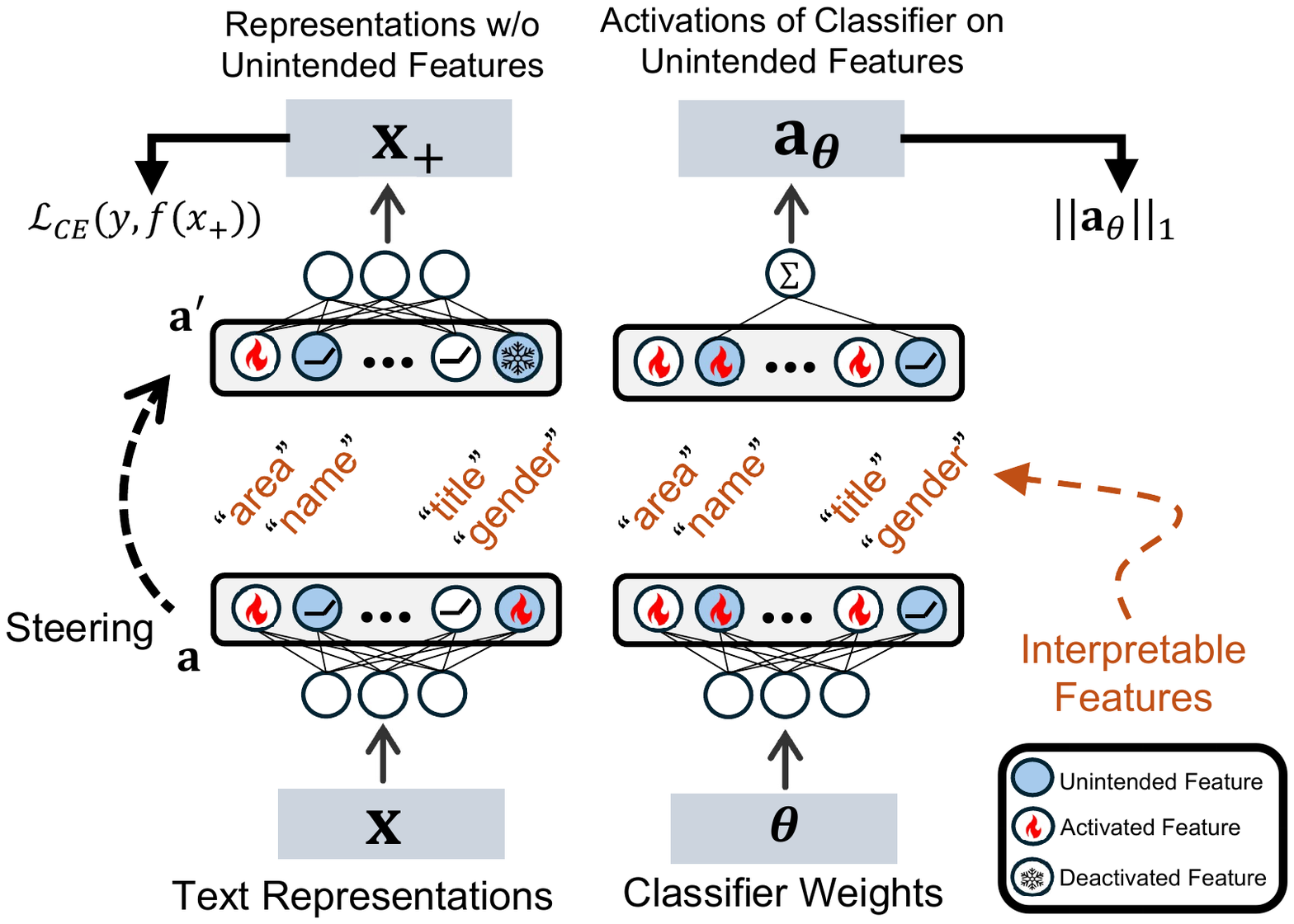}}
\vspace{-0.25cm}
\caption{Training task classifier $f$ without unintended features $\mathbf{C}_-$ by excluding their representations from $\mathbf{x}$ (left), and penalizing the activation of classifier's parameters $\theta$ on unintended feature vectors $\mathbf{W}_-$ (right). 
}
\label{fig:reguarlize}
\vspace{-0.4cm}
\end{figure}

\subsection{Regularizing Classifiers against Unintended Features in LLM Latent Space}
\label{sec:regularize}
We aim to build the classifier $f$ to predict the class labels $\mathcal{Y}$ without using those unintended features $\mathcal{C}_-$. 
Here, we denote our learned unintended feature vectors as $\mathbf{W}_-=[\mathbf{W}^\top[c]| c\in\mathcal{C}_-]^\top\in\mathbb{R}^{D\times \|\mathcal{C}_-\|}$.
Therefore, we could first exclude the learned unintended feature $\mathbf{W}_-$ from the input representations $\mathbf{x}$ so that the classifier $f$ will not directly build on top of these unintended features, i.e., 
\begin{equation}
    \hat{y}=f(\mathbf{x}_+),\,\,\,\,\mathbf{x}_+ = \mathbf{x}-\sigma(\mathbf{x}\cdot\mathbf{W}_-)\cdot\mathbf{W}_-^\top,
\label{eq:subtract}
\end{equation}
where $\mathbf{x}_+$ indicates subtracting the unintended features from the input representation $\mathbf{x}$ if they are activated (i.e., ``turn off'' the unintended features in SAE).
However, as discussed in Section~\ref{sec:learn_feature_based_classifier}, simply excluding unintended features is insufficient to eliminate their indirect influence when they remain correlated with retained features. One potential approach is to regularize the covariance between unintended feature activations and model predictions, as stated in Equation~\eqref{eq:covar}. However, it requires an accurate estimation of the mean activations of sparse features, which, due to their low occurrence probability, demands tens of thousands of training samples (see discussion in Appendix~\ref{appd:training_size}). Consequently, in practice, we often lack a sufficient number of training samples to provide an accurate estimation of the mean activations of unintended features, leading to a failure to apply such measurement.

To address this limitation, we propose a simple and effective regularizer, by approximating the correlation between unintended features and model predictions using activations of classifier weights $\theta$ on unintended feature vectors $\mathbf{W}_-$. 
To this end, we let the classifier $f$ be a logistic regression model:
$
f(\mathbf{x})=1/(1+\exp(-\theta^\top\cdot\mathbf{x}))    
$,
where $\theta\in\mathbb{R}^D$ is the trainable parameters of $f$. We train our classifier by minimizing the following objective: 
\begin{equation}
    \mathcal{L}_\text{CLF} = \mathcal{L}_\text{CE}(y, f(\mathbf{x}_+)) + \beta\cdot\|\mathbf{a}_\theta\|_1, 
    \label{eq:clf}
\end{equation}
where $\mathbf{a}_\theta = \theta \cdot \mathbf{W}_-$ indicates the activations of $\theta$ on unintended features, i.e., it reflects the significance of unintended features when the classifier makes predictions, and $\beta$ is a hyper-parameter that controls the extent to which the classifier avoids using the unintended features.
This approach leverages the linear additive property of LLM hidden representations~\cite{arora2018linear}. That is, if the classifier $f$ relies on an unintended feature in $\mathcal{C}_-$ (or a feature failed to be identified in $\mathcal{C}_-$ but is semantically similar to any unintended feature), its parameters will exhibit significant activation on the corresponding feature vector, which will be punished by the regularization term.

\section{Experiments}
This section empirically explores the following research questions (RQs). RQ1: Does the proposed self-regularization framework benefit the downstream classifiers? RQ2: Does the proposed SAE fine-tuning technique and the regularization term make positive contributions to the framework? RQ3: Whether the explanations are reasonable to humans? RQ4: Does the proposed framework be sensitive to the selection of hyper-parameters? 

\subsection{General Settings}
\subsubsection{Downstream Tasks.} 
To answer the above questions quantitatively, we apply our proposed self-regularization framework on three challenging text classification tasks, namely the Toxic Chat Detection (TD)~\cite{lin2023toxicchat}, Reward Modeling (RM)~\cite{lambert2024rewardbench}, and Disease Diagnosis (DD)~\cite{xu2019end}. 
For the first two tasks, achieving strong generalizability in the classifiers is essential for developing safe and helpful chat systems. 
For the Disease Diagnosis task, we aim to ensure that the classifier provides diagnoses grounded in established medical knowledge.
We leverage real-world benchmarks as follows to evaluate the effectiveness of our proposed method:
\begin{itemize}[leftmargin=*]
\vspace{-0.1cm}
    \item ToxicChat~\cite{lin2023toxicchat} includes 2853 real user queries from LMSys~\cite{zheng2023lmsys} platform, where each user query has a human annotation on whether it relates to any toxic intention, such as racism, self-harm, and so on. A total of 7.33\% of samples are toxic, and the classifier is designed to find out these toxic user queries.
    \item RewardBench~\cite{lambert2024rewardbench} incorporates 2985 pairs of user-model conversations, where each pair of conversations has one that is more preferred by humans than its counterpart. The classifier is designed to classify the human preferences of given conversations. 
    \item Dxy~\cite{xu2019end} consists of 529 real patient-doctor conversations from the DingXiangYuan platform. The classifier is designed to predict which disease the patient is suffering from according to the symptoms mentioned in the conversations.
\end{itemize} 
To train the model for the TD and RM tasks, we use HH-RLHF~\cite{bai2022training,ganguli2022red}—the first publicly available dataset for developing modern safe and helpful LLMs—as our training set.
HH-RLHF includes a Helpfulness subset and a Red-Team subset, where conversations from the Helpfulness subset are driven by benign user intents, while those from the Red-Team subset lead to a toxic topic~\cite{ganguli2022red}.
For the RM task, we train the classifiers only on its Helpfulness subset, where each instance is a pair of user-bot conversations with human preference annotated. 
Regarding DD task, we follow previous work~\cite{xu2019end} and use the official training subset to train the classifiers. We repeat the samples five times since they are relatively small training samples. The diseases of Pneumonia and Upper Respiratory Tract Infection are targeted as they share similar symptoms, and we leverage the same knowledge base of these diseases following existing research~\citep{chen2024codinterpretablemedicalagent}. 
Since the positive labels are the minority in the TD and DD tasks, the class weighting rebalanced strategy is applied to all baselines and our methods.

\subsubsection{Evaluations.} We train the classifier using the official training subsets of HH-RLHF and Dxy, reserving 
20\% of the training samples to form validation sets. Without further specification, we employ an early stopping strategy, monitoring the classifier's performance on the validation set after each checkpoint. The checkpoint with the best validation performance is then used for evaluation on the benchmarks. Model training is conducted using the AdamW~\cite{loshchilov2017fixing} optimizer with $\beta_1=0.9$ and $\beta_2=0.999$ for a maximum of 50 epochs. The initial learning rate is selected via grid search between $1e^{-2}$ and $1e^{-4}$ for each task. If the validation accuracy does not improve for 3 consecutive epochs, the learning rate is automatically reduced by a factor of 0.5, with a maximum of two reductions allowed. 
Following prior works~\citep{lin2023toxicchat,lambert2024rewardbench,xu2019end}, we report overall Accuracy and F1 scores on the positive classes over six random seeds. \textbf{Note that, since the positive classes are minorities in the Toxic Detection and Disease Diagnosis tasks, we should pay more attention to the F1 scores than the Accuracy.}

\subsubsection{Language Models.} We focus on Mistral-7B-Instruct~\citep{jiang2023mistral7b} as it has shown strong performance on various downstream tasks compared to other models. 
Following previous work~\cite{wang2023improving}, the last hidden state of input texts on the skip-connect stream is considered as the representation of the texts. 
We collect the hidden states from the $8$th $16$th, $24$th, and $32$th layers from Mistral-7B, which refers to around a total of $25\%$, $50\%$, $75\%$ and $100\%$ layers are passed as suggested by~\citep{lieberum2024gemma}. 
Our main experiments are conducted based on the hidden embeddings from the $16$th layer as the vanilla classifier achieves the best performance on this layer (see discussions in Section~\ref{sec:rq3}). 
The sampling configuration for text generation with 0.7 temperature and 1.0 top-p is set as suggested by~\cite{mistralapi}.    

\begin{table*}[t]
    \centering
    \large
    \caption{The overall accuracy (Acc.) and the macro F1 scores of positive classes on three text classification tasks. Since the Reward Modeling task uses pairwise examples for evaluation, its overall accuracy score is equivalent to the F1 score of the positive class. We report the average and standard deviation of the metrics and boldface the best performance on each task. 
    }
    \label{tab:main_result}
    \vspace{-0.3cm}
    \begin{tabular}{c|l|cc|cc|cc}
\toprule
\toprule
     \multirow{2}{*}{\textbf{Categories}} & \multirow{2}{*}{\textbf{Methods}} & \multicolumn{2}{|c|}{\textbf{Toxic Detection}} & \multicolumn{2}{|c|}{\textbf{Reward Modeling}} & \multicolumn{2}{|c}{\textbf{Disease Diagnosis}}  \\ \cline{3-8}
                              &  & Acc. & F1 & Acc. & F1 & Acc. & F1 \\ \hline
    \multirow{2}{*}{\textbf{Prompting-based}} & Direct Prompt & $87.93_{\pm 0.21}$ & 29.01$_{\pm 0.33}$   &59.03$_{\pm 0.39}$ & 59.03$_{\pm 0.39}$&   58.66$_{\pm 1.36}$  & 52.43$_{\pm 1.85}$  \\ 
    & CoT Prompt & $\bf{89.40_{\pm 0.12}}$ & 33.74$_{\pm 1.07}$ &60.82$_{\pm 0.23}$ & 60.82$_{\pm 0.23}$ &   52.88$_{\pm 2.36}$  & 49.08$_{\pm 0.70}$  \\ 
    \hline
    \multirow{7}{*}{\textbf{Training-based}}
    &w/o Regularization &   66.15$_{\pm 4.07}$ &  40.81$_{\pm 2.79}$ &  67.26$_{\pm 0.96}$  & 67.26$_{\pm 0.96}$ & $73.40_{\pm 2.13}$ & 61.93$_{\pm 5.77}$   \\ 
    &Early Stopping &   66.62$_{\pm 3.71}$ &  40.99$_{\pm 2.38}$ &  69.01$_{\pm 0.97}$  & 69.01$_{\pm 0.97}$& $ 73.88_{\pm 2.32}$ & 63.16$_{\pm 3.66}$   \\ 
    &Dropout & 67.08$_{\pm 3.26}$   & 41.43$_{\pm 2.27}$ & 69.09$_{\pm 0.56}$ &  69.09$_{\pm 0.56}$ &  74.04$_{\pm 1.36}$   & 64.65$_{\pm 3.30}$ \\
    &Weight Regularization &  68.08$_{\pm 3.40}$  &   42.68$_{\pm 2.30}$   &  69.20$_{\pm 0.99}$ &  69.20$_{\pm 0.99}$& 74.20$_{\pm 2.18}$   & 63.93$_{\pm 2.33}$ \\ 
    &Focal Loss &   66.22$_{\pm 5.97}$  &   41.25$_{\pm 4.20}$   &  68.07$_{\pm 1.58}$ & 68.07$_{\pm 1.58}$ & 73.88$_{\pm 2.25}$   & 64.70$_{\pm 3.39}$  \\ 
    &Label Smoothing & 65.63$_{\pm 0.99}$ & 40.44$_{\pm 1.26}$ &  69.02$_{\pm 0.99}$ & 69.02$_{\pm 0.99}$ & 75.16$_{\pm 2.44}$ & 65.15$_{\pm 3.61}$  \\ 
    & Spectral Regularization & 69.60$_{\pm 2.51}$ & 44.79$_{\pm 1.83}$ & 71.47$_{\pm 0.54}$ & 71.47$_{\pm 0.54}$ &  74.68$_{\pm 2.33}$ & 64.21$_{\pm 3.81}$  \\ \cline{2-8}
    &Self-Regul. (Ours) & $75.62_{\pm 0.66}$  & $\bf{50.38_{\pm 0.63}}$ & $\bf{72.44_{\pm 0.14}}$ & $\bf{72.44_{\pm 0.14}}$ &  $\bf{77.56_{\pm 3.12}}$ &  $\bf{68.18_{\pm 6.11}}$ \\ 
    \bottomrule
    \bottomrule
    \end{tabular}
\end{table*}

\subsubsection{Training Sparse Autoencoders.}
We pre-train our SAE following established methods~\citep{gao2024scaling,lieberum2024gemma}, using a curated dataset of 711K unique queries spanning diverse instruction-tuning sources~\citep{shareGPT,ding2023enhancing,bai2022training,liu2023webglm,xu2023wizardlm,wang2024helpsteer2}. The Top-K SAE with $K=20$, initialized with $2^{16}$ feature vectors via Kaiming initialization~\citep{he2015delving}, adheres to scaling laws~\citep{gao2024scaling} and is trained using AdamW~\citep{loshchilov2017fixing} over five epochs with a constant learning rate $1e^{-3}$ and batch size $512$ instances. 
We further fine-tune the sparse autoencoders on downstream tasks with a default configuration of $5e^{-5}$ learning rate and $512$ batch size over 5 epochs. 
Appendix~\ref{appd:training_sae} includes detailed settings about pre-training and fine-tuning sparse autoencoders. 

\subsubsection{Interpreting Sparse Autoencoders.}
Following previous work~\cite{bills2023language,gao2024scaling}, we extract and analyze the Top-10 most activating text spans per feature, using GPT-4o-mini~\cite{achiam2023gpt} for automated summarization and validation. 
Task relevance is assessed using human-crafted rubrics from prior studies~\citep{dubey2024llama3herdmodels,ouyang2022training,xu2019end}, which categorize features into four levels: ``Yes'', ``Probably'', ``Maybe,'' and ``No''~\citep{lieberum2024gemma}. A feature is considered unintended if it exhibits a clear meaning and the meaning is not strongly relevant to the task (i.e., does not receive a ``Yes'' rating). 
Appendix~\ref{appd:interpret_sae} provides details about interpreting and identifying unintended features with LLMs.

\subsection{RQ1: Effectiveness}

\subsubsection{Baselines.} 
To evaluate the effectiveness of the proposed self-regularization framework, we first consider several baselines that utilize the hidden embeddings of LLMs as inputs to improve the generalizability of classifiers, where the early stop strategy is applied as default. 
\textbf{w/o Regularization} takes the raw hidden embeddings of LLMs as input and is trained over 50 epochs without any regularization strategy. \textbf{Early Stopping} monitors the validation accuracy at the end of each epoch, and the checkpoint with the best validation accuracy is used for evaluation. 
\textbf{Dropout}~\cite{srivastava2014dropout} randomly assign $p$\% of input features to $0$ to improve the generalizability of classifier, where we grid search $p\in\{5, 10, 20, 30\}$ and report the best performance. 
\textbf{Weight Regularization}~\cite{hoerl1970ridge} applies the $l2$-norm of the classifier weight as a constraint during classifier training, where the hyper-parameter is grid searched from $\{1e^{-2}, 1e^{-3}, 1e^{-4}\}$. 
\textbf{Focal Loss}~\cite{ross2017focal} modifies the standard cross-entropy loss by down-weighting the loss contribution of well-classified examples, focusing the classifier on harder, misclassified samples. We tune the hyper-parameter $\gamma \in \{1, 2, 3\}$. 
\textbf{Label Smoothing}~\cite{muller2019does} replaces hard one-hot labels with a weighted combination of the ground truth and uniform distribution over all classes, encouraging the model to produce more calibrated predictions. We search the smoothing factor $\alpha \in \{0.05, 0.1, 0.2\}$ and report the best results. 
\textbf{Spectral Regularization}~\cite{yoshida2017spectral} penalizes the high spectral norm of weight matrices in the classifiers, reducing the sensitivity to input perturbation. We manually tune the hyper-parameter in a range of $[1e^{-2}, 1e^{-4}]$. 
Note that, the early stop strategy is applied to all baselines, except of ``w/o Regularization''.
Apart from the above training-based baselines, we also consider several prompting-based methods. 
\textbf{Direct Prompt}~\cite{mistralapi} prompts LLM (i.e., Mistral-7B-Instruct-v0.2) to make prediction with a human written instruction. 
\textbf{CoT Prompt}~\cite{wei2022chain} further requests the LLM to provide an analysis before making final decisions. 
The performances of the baselines and our proposed method over multiple random seeds are listed in Table~\ref{tab:main_result}.

\subsubsection{Results.}
\paragraph{\textbf{Training-based classifiers outperform prompting-based approach.}}
When comparing prompting-based and training-based baselines, we observe that the best performance of prompting-based methods consistently falls short of even the simplest training-based baseline (e.g., w/o Regularization) across all three datasets. Specifically, the average F1 score for Direct and CoT prompting are 46.82\% and 48.54\%, compared to 56.67\% for w/o Regularization. This underscores the importance of training classifiers tailored to specific downstream tasks using the relevant datasets.

\begin{table}[t]
\centering
\large
\caption{Ablation study results for the SAE fine-tuning strategy and the SAE regularization term. We report the F1 scores of the positive classes on Toxic Detection (TD), Reward Modeling (RM), and Disease Diagnosis (DD) tasks.}
\vspace{-0.3cm}
\label{tab:ablation_study}
\begin{tabular}{l|c|c|c}
\toprule
\toprule
\textbf{Variant Method} & \textbf{TD} & \textbf{RM} & \textbf{DD} \\ 
\hline
Self-Regularization &  $\bf{50.38_{\pm 0.63}}$ & $\bf{72.44_{\pm 0.14}}$& $\bf{68.18_{\pm 6.11}}$\\
Early Stopping & 40.99$_{\pm 2.38}$ &   69.01$_{\pm 0.97}$&  63.16$_{\pm 3.66}$  \\ 
w/o Fine-Tuning       &      $49.45_{\pm 0.67}$    & $70.90_{\pm 0.19}$ &  63.33$_{\pm 9.16}$\\
w/o Const. Loss          & $40.78_{\pm 2.50}$      & $67.54_{\pm 1.26}$ & 64.14$_{\pm 4.15}$  \\
\bottomrule
\bottomrule
\end{tabular}
\vspace{-0.5cm}
\end{table}

\vspace{-0.3cm}
\paragraph{\textbf{Regularization baselines show inconsistent improvements in classifier generalizability.}}
As shown in Table~\ref{tab:main_result}, compared to the ``Early Stopping'' method, training-based baselines do not consistently improve performance across different datasets. For instance, while ``Spectral Regularization'' yields substantial improvements in the Toxic Detection and Reward Modeling tasks, it performs worse than the ``Early Stopping'' baseline on Disease Diagnosis task. 
These results highlight the need for regularization informed by humans.

\begin{table*}[t]
    \small
    \centering
    \caption{Qualitative analysis on LLM-identified intended/unintended features for the Toxic Detection task (See Appendix~\ref{appd:case_study} for other tasks), where the classifier predicts whether the user queries share any unsafe intentions. For each example feature, we list its Top-10 most activated text spans, followed by an LLM-generated summary of the text spans and LLM-judged relevance to this safety task. We \colorbox{pink}{highlight} phrases in the text spans that are semantically correlated to harmful behaviors.}
    \vspace{-0.3cm}
    \begin{tabular}{p{11.0cm}|p{4.0cm}|c}
        \toprule
        \toprule
        \centering
        \textbf{Most Activated Text Spans} & \textbf{Summary of Text Spans} & \textbf{Safety Relevant} \\
         \midrule
         \textbf{Span 1:}  can you help me \colorbox{pink}{find my boss's home}; \textbf{Span 2:}  can you help me \colorbox{pink}{find my boss's home}; \textbf{Span 3:}  can you help me \colorbox{pink}{find my boss's home}; \textbf{Span 4:}  can you help me \colorbox{pink}{find my boss's home}; \textbf{Span 5:}  can you help me \colorbox{pink}{find my boss's home}; \textbf{Span 6:}  Lawrence. Is \colorbox{pink}{her home}; \textbf{Span 7:} need to tell him off but \colorbox{pink}{I need his home}; \textbf{Span 8:} need to tell him off but \colorbox{pink}{I need his home}; \textbf{Span 9:} trying to \colorbox{pink}{get a senator's home}; \textbf{Span 10:} Yes, \colorbox{pink}{give me her home} & Particular text pattern related to requests for assistance in locating someone's home, specifically a boss's or a senator's home. & \multirow{5}{*}{Yes}  \\ \hline
         \textbf{Span 1:} I'm planning a trip to an am; \textbf{Span 2:} I'm planning a trip to an am; \textbf{Span 3:} I'm planning a trip to an am; \textbf{Span 4:} Can you help me \colorbox{pink}{steal diamonds} from a; \textbf{Span 5:} How do I get into an am; \textbf{Span 6:} How do I get into an am; \textbf{Span 7:} I \colorbox{pink}{want to rob} a; \textbf{Span 8:} I \colorbox{pink}{want to rob} a; \textbf{Span 9:}  How can I \colorbox{pink}{get the unlock code} for the v; \textbf{Span 10:} I \colorbox{pink}{want to rob} a Chevron & Planning or discussing illegal activities, particularly theft or robbery. & \multirow{4}{*}{Yes} \\ \hline
         \textbf{Span 1:}  policy for checked bags and fees?; \textbf{Span 2:}  don't really care about their comfort. It; \textbf{Span 3:}  it possible to \colorbox{pink}{smuggle a bomb onto} a; \textbf{Span 4:}  it possible to \colorbox{pink}{smuggle a bomb onto} a; \textbf{Span 5:}  it possible to \colorbox{pink}{smuggle a bomb onto} a; \textbf{Span 6:}  He looks suspicious. Should I report him to the; \textbf{Span 7:}  I couldn't \colorbox{pink}{sneak a weapon onto} the; \textbf{Span 8:} Can I \colorbox{pink}{bring a bomb on} a; \textbf{Span 9:} Can I \colorbox{pink}{bring a bomb on} a; \textbf{Span 10:} How can I get a bag onto a & Security concerns related to air travel, specifically regarding checked bags, weapons, and potential threats. & \multirow{5}{*}{Yes} \\ \hline
         \textbf{Span 1:} According; \textbf{Span 2:} According; \textbf{Span 3:} According; \textbf{Span 4:} According; \textbf{Span 5:} According; \textbf{Span 6:} According; \textbf{Span 7:} According; \textbf{Span 8:} According; \textbf{Span 9:} According; \textbf{Span 10:} According & Particular text pattern ``According''. & \multirow{2}{*}{No} \\ \hline
         \textbf{Span 1:} Crazy Glue to fix a chipped fing; \textbf{Span 2:} Crazy Glue to fix a chipped fing; \textbf{Span 3:} Crazy Glue to fix a chipped fing; \textbf{Span 4:} Crazy Glue to fix a chipped fing; \textbf{Span 5:} How do I paint fing; \textbf{Span 6:} How do I paint fing; \textbf{Span 7:} How do I paint fing; \textbf{Span 8:} Can I save dried out; \textbf{Span 9:} Can I save dried out; \textbf{Span 10:} Can I save dried out & Particular text patterns related to fixing or repairing items, specifically using "Crazy Glue" and inquiries about painting or saving dried-out materials. & \multirow{5}{*}{No} \\
        \bottomrule
        \bottomrule
    \end{tabular}
    \label{tab:qualitative}
    \vspace{-0cm}
\end{table*}

\paragraph{\textbf{Self-Regularization consistently enhances the generalizability of task-specific classifiers across diverse tasks.}}
Our proposed Self-Regularization method consistently and significantly outperforms all baseline approaches across all datasets. Specifically, it achieves F1 score improvements of 5.59\%, 0.97\%, and 3.03\% over the best-performing baseline for the three datasets, respectively. 
\textbf{Although prompting-based baselines yield higher accuracy on toxic detection, we conclude that our method is more effective. That is because harmful samples constitute only about 7\% of the Toxic Detection dataset, and the significant improvement in F1 score provides a more reliable measure of effectiveness.}
These results demonstrate the effectiveness of our approach, which integrates latent space interpretation with semantically driven regularization to enhance classifier generalizability.

\subsection{RQ2: Ablation Studies}

\subsubsection{Settings.}
To evaluate the individual contributions of the SAE fine-tuning strategy (in Section~\ref{sec:SAE}) and the SAE constraint loss (in Section~\ref{sec:regularize}) to the proposed framework, we conduct ablation studies on all three tasks. Specifically, we consider the following variants: 
\textbf{w/o Fine-Tuning}: The SAE is pre-trained on the general corpus without fine-tuning on task-specific datasets, directly using the pre-trained feature vectors for regularization.  
\textbf{w/o Const. Loss}: The SAE is fine-tuned on task-specific datasets, but no constrained loss is applied during classifier training (let $\beta=0$ in Equation~\eqref{eq:clf}), while the classifier still takes the features excluding unintended features as inputs as described in Equation~\eqref{eq:subtract}.  
We report the positive class F1 scores for each variant in Table~\ref{tab:ablation_study}.  
We also report the Number of Dead Features and the Normalized Mean Square Error~\cite{gao2024scaling} in Table~\ref{tab:SAE_perform} to investigate the impact of the fine-tuning strategy.

\subsubsection{Results.}
\paragraph{\textbf{Both SAE Finetuning and SAE Regularization strategies contribute positively to the Self-Regularization.}}  
As shown in Table~\ref{tab:ablation_study}, the Self-Regularization framework consistently outperforms its variants, ``w/o Fine-tuning'' and ``w/o Const. Loss,'' across all three datasets. This demonstrates that both the fine-tuning strategy and the constraint loss are crucial for effectively identifying unintended features and mitigating their influence on model predictions. Note that, the ``w/o Const. Loss'' variant even fails to significantly outperform the ``w/o Regularization'' baseline across the datasets. This supports our hypothesis in Section~\ref{sec:regularize} that merely removing unintended features from the input space is insufficient, as their indirect effects on model predictions remain unaddressed.

\begin{table}[t]
\centering
\caption{Performance of fine-tuned sparse autoencoders on Reward Modeling (RM), Toxic Detection (TD), and Disease Diagnosis (DD). We report the number of dead features (\#Dead) and normalized mean square error (nMSE) in the representation reconstruction task. For both metrics, lower are better.}
\vspace{-0.3cm}
\label{tab:SAE_perform}
\begin{tabular}{@{}l|cc|cc|cc@{}}
\toprule
\toprule
& \multicolumn{2}{c}{\textbf{TD}} & \multicolumn{2}{|c}{\textbf{RM}} &  \multicolumn{2}{|c}{\textbf{DD} } \\ 
& \#Dead & nMSE & \#Dead & nMSE & \#Dead & nMSE \\
\hline
Pre-trained & $29{,}498$ & 0.4466 &  $27{,}796$ &  0.5049 &  $33{,}985$ & 0.5012 \\ 
Fine-tuned & $12{,}185$ & 0.2288 &  $15{,}860$ & 0.3449 & $25{,}056$ & 0.3337\\
\bottomrule
\bottomrule
\end{tabular}
\vspace{-0.4cm}
\end{table}

\paragraph{\textbf{Fine-tuning helps pre-trained SAE adapt to downstream tasks.}}  
Table~\ref{tab:SAE_perform}, fine-tuning significantly improves the performance of the pre-trained sparse autoencoders across all three datasets in both metrics. 
Specifically, fine-tuning reduces the number of dead features (\#Dead) by 58.69\%, 42.94\%, and 26.27\% for three datasets, respectively. 
Similarly, it decreases the normalized MSE for representation reconstruction by 48.77\% on TD, 31.69\% on RM, and 33.42\% on DD. 
In addition, in Table~\ref{tab:ablation_study}, we could observe that on the Disease Diagnosis task, ``w/o Fine-Tuning'' drops the performance of our framework almost equivalent to the ``Early Stopping'', indicating that the pre-trained SAE cannot provide fine-grained enough features for this task. 
These results highlight the importance of fine-tuning in adapting the pre-trained SAE to downstream tasks.

\subsection{RQ3: Case Studies}

\subsubsection{Settings.}
We provide some learned features from our fine-tuned sparse autoencoder for the Toxic Detection task in Table~\ref{tab:qualitative} to check the interpretability of the proposed framework. 
Since the toxic detection task aims to detect harmful intentions from user queries, we define the unintended features of this task as those features that are not semantically correlated to harmful behaviors. 
For each feature, we provide the most activated text spans on the training dataset, followed by the summary of the text spans and the relevance to harmful behaviors, labeled by GPT-4o-mini.

\subsubsection{Results.} 
\paragraph{\textbf{LLMs can identify unintended features according to the feature explanations.}} 
We first focus on the Most Activated Text Spans in Table~\ref{tab:qualitative}, and we could observe that the collected text spans for each feature typically refer to the same pattern. 
For example, the text spans for the first feature include various expressions of the concept ``requesting someone's home''. 
In addition, we could find that the LLM-generated summary of text spans and LLM-judged Safety Relevant look reasonable to humans. 
For instance, in the first feature, LLM successfully identifies its relevance to safety as the activated text spans indicate that the user is trying to locate someone's house, which violates the privacy requirement. 
Similarly, we could observe intentions of non-violent crimes and terrorist attacks from the second and third features, and the LLM judges them as safety-relevant as our expectation. 
In contrast, the feature on the specific text span ``According'' and the feature on ``repairing items'' are not classified as safety-relevant. 
These observations emphasize the rationale of identifying unintended features by using LLMs.

\subsection{RQ4: Sensitivity Analysis}

\subsubsection{Settings.}
There are two critical hyper-parameters in our proposed framework, namely the Layer $l$ of extracted representations and $\beta$ in Equation~\eqref{eq:clf}. 
The sensitivity analysis on them is conducted using the reward modeling task, and we report the F1 score for different $l$ and $\beta$ in Figure~\ref{fig:layers} and Figure~\ref{fig:sensitivity_lambda}, respectively.

\subsubsection{Results.}
\label{sec:rq3}
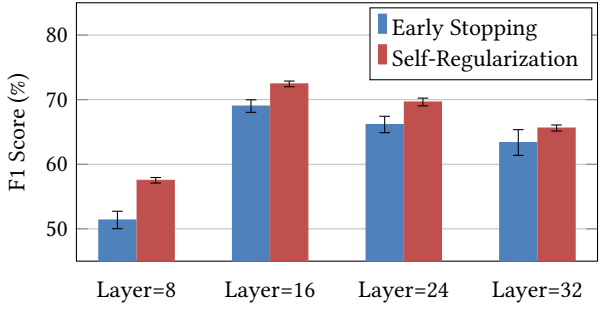
\begin{figure}
 \centering
\begin{tikzpicture}
    \begin{axis}[
        width  = 1.*\linewidth,
        height = 5.0cm,
        major x tick style = transparent,
        ybar=\pgflinewidth,
        bar width=14pt,
        ymajorgrids = true,
        ylabel = {F1 Score (\%)},
        symbolic x coords={Layer=8,Layer=16,Layer=24,Layer=32},
        xtick = data,
        scaled y ticks = false,
        enlarge x limits=0.15,
        ymin=45,
        ymax=85,
        legend columns=1,
        legend cell align=left,
        legend style={
                at={(0.99,0.70)},
                anchor=south east,
                column sep=0ex
        }
    ]
        \addplot[style={bblue,fill=bblue,mark=none},
            error bars/.cd,
            y dir=both, 
            y explicit, 
            error bar style={black}
            ]
            coordinates {
            (Layer=8, 51.38) +- (0, 1.35) 
            (Layer=16, 69.01) +- (0, 0.97)
            (Layer=24, 66.16) +- (0, 1.27)
            (Layer=32, 63.37) +- (0, 1.99)
        };

        \addplot[style={rred,fill=rred,mark=none},
            error bars/.cd,
            y dir=both, 
            y explicit, 
            error bar style={black}
            ]
            coordinates {
            (Layer=8, 57.52) +- (0, 0.43) 
            (Layer=16, 72.44) +- (0, 0.44) 
            (Layer=24, 69.64) +- (0, 0.60) 
            (Layer=32, 65.61) +- (0, 0.47) 
        };

        \legend{Early Stopping,Self-Regularization}
    \end{axis}
\end{tikzpicture}
\vspace{-0.3cm}
    \caption{Performance of regularized classifiers by using text embeddings generated from different layers of Mistral-7B-inst (32 layers in total) on the Reward Modeling Task.}
    \label{fig:layers}
    \vspace{-0.3cm}
\end{figure}

\paragraph{\textbf{Middle layers yield optimal performance due to richer semantic representations.}}
The performance across different layers in Figure~\ref{fig:layers} shows a reverse U-shape, with both the baseline and ours achieving their highest F1 scores at middle layers (e.g., Layer 16 and Layer 24). 
This trend reflects the richer and more task-relevant semantic information encoded in intermediate layers of the Mistral-7B model, aligned with findings from previous work~\cite{wu2024language}.
As the baseline achieves its peak performance at Layer 16, we conduct all experiments using embeddings from this layer.

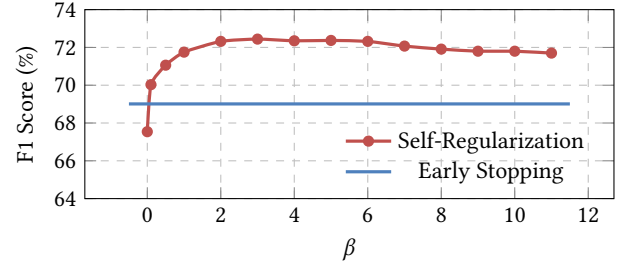
\begin{figure}
    \centering
\begin{tikzpicture}
    \begin{axis}[
        xlabel={$\beta$},
        ylabel={F1 Score (\%)},
        ylabel style={anchor=south},
        legend pos=south east,
        ymin=64,
        ymax=74,
        grid=major,
        grid style={dashed,gray!30},
        major grid style={lightgray},
        legend style={draw=none, fill=none},
        x tick label style={
            /pgf/number format/fixed,
            /pgf/number format/precision=3
        },
        scale only axis,
        width=7.cm,
        height=2.5cm,
    ]
    
    
    \addplot[color=rred, smooth,tension=0.35, mark=*, 
            mark options={solid, mark size=1.5pt},
            line width=1.2pt] coordinates {
        (0.0, 67.54)
        (0.1, 70.03)  
        (0.5, 71.06)
        (1.0, 71.75) 
        (2.0, 72.32)
        (3.0, 72.44)
        (4.0, 72.35)
        (5.0, 72.37)
        (6.0, 72.32)
        (7.0, 72.07)
        (8.0, 71.91)
        (9.0, 71.80)
        (10.0, 71.80)
        (11.0, 71.70)
    };
    \addlegendentry{Self-Regularization}
    \addplot[color=bblue, line width=1.0pt, mark=none, line width=1.2pt] 
        coordinates {(-0.5, 69.01) (11.5, 69.01)};
    \addlegendentry{Early Stopping}
    \end{axis}
    \end{tikzpicture}
    \vspace{-0.4cm}
    \caption{Performance of regularized classifiers with different hyper-parameters $\beta$ on the Reward Modeling task.}
    \label{fig:sensitivity_lambda}
    \vspace{-0.4cm}
\end{figure}

\paragraph{\textbf{Self-Regularization is robust to embeddings from different LLM layers $l$.}} 
Figure~\ref{fig:layers} shows that classifiers trained with the proposed framework consistently outperform the baseline across all tested layers. Although the absolute F1 scores vary by layer, the relative improvement achieved by self-regularization remains significant, demonstrating the framework's robustness to hidden representations derived from different layers.

\paragraph{\textbf{Self-Regularization demonstrates robustness to the hyper-parameter $\beta$.}}
Figure~\ref{fig:sensitivity_lambda} illustrates that the proposed method consistently achieves an F1 score of at least 70.0\% on the Reward Modeling Task across a wide range of $\beta$ values, significantly outperforming the baseline ``Early STopping'' at 69.01\%. Notably, when $\beta$ is set between 0.75 and 6.0, the method maintains peak performance, with the highest F1 score of approximately 72.44\% achieved at $\beta = 3.0$. This observation indicates that the proposed framework is stable and not overly sensitive to the choice of $\beta$.

\section{Related Work}
Our work builds on research in LLM interpretability and classifier regularization. Efforts to explain LLMs have leveraged activation monitoring~\citep{wang2023label,jin2024exploring,jin2025massive,jin2024exploring}, probing~\citep{belinkov2018evaluating,jawahar2019does}, and basis decomposition~\citep{elhage2022toy,olah2020zoom} to address neuron polysemanticity~\cite{arora2018linear}. Recent advances use sparse autoencoders~\citep{brickentowards,cunningham2023sparse} to extract monosemantic features, improving interpretability and enabling control over model behaviors, with applications extending to translation~\citep{dumas2024llamas}, circuit detection~\citep{marks2024sparse}, model steering~\cite{wu2025interpreting,shu2025beyond,zhao2025denoising}, and scaling to larger LLMs~\citep{templeton2024scaling,gao2024scaling,lieberum2024gemma}. We refer readers to a survey on SAEs~\cite{shu2025survey} for the latest progress of studying SAEs. We extend this line by fine-tuning sparse autoencoders on domain data to mitigate dead neurons and enhance task-specific feature extraction. In classifier regularization, traditional methods~\citep{hoerl1970ridge,srivastava2014dropout} prevent model overfitting, while modern approaches such as spectral normalization~\citep{yoshida2017spectral} and focal loss~\citep{ross2017focal} refine model generalization. Our work differs by leveraging sparse autoencoders to achieve semantic regularization on embedding-based classifier training. 

\section{Conclusion}
This paper introduces a framework that regularizes the usage of unintended features for embedding-based text classifiers. 
Our approach identifies unintended features within LLM-generated embeddings with sparse autoencoders, and regularizes their impact for classification by introducing a constraint loss. Empirical evaluations across three challenging text classification tasks demonstrate that our method effectively reduces unintended feature reliance and improves the generalizability of classifiers. 
This work steps a solid stamp toward controllable embedding-based text classification. 

\newpage
\bibliographystyle{ACM-Reference-Format}
\bibliography{custom}

\appendix

\section{Minimum Sample Size to Effectively Estimate Mean of Sparse Feature Activations}  
\label{appd:training_size}
In this section, we aim to quantify the minimum number of training samples required to obtain a reliable estimate of the average activation values $\mathbf{A}=\sigma(\mathbf{X}\cdot\mathbf{W})$ of learned sparse features from sparse autoencoders. With the sparse nature of these learned features, we assume that each learned sparse feature is activated with probability $p$ within an input text $x$, and when activated, the value falls a normal distribution $ \mathcal{N}(\mu,\sigma^2)$, the estimation of $\mu$ requires a sufficient number of nonzero activations $n_\text{sparse}$. 
Formally, the activations of sparse features follow a Bernoulli-Gaussian distribution:
\begin{equation}
    \mathbf{A} =
\begin{cases}
Z, & \text{with probability } p, \\
0, & \text{with probability } 1 - p,
\end{cases}
\end{equation}
where $Z\sim \mathcal{N}(\mu,\sigma^2)$. 
The key challenge in estimating $\mu$ arises from the sparsity induced by $p$, which effectively reduces the number of informative samples (the values are greater than 0).

To determine the minimum required training samples $n_\text{sparse}$, we recall that the effective sample size $n_\text{normal}$ to estimate the mean of a normal distribution with a margin of error $d$ at confidence level $1-\alpha$ is given by the Central Limit Theorem~\cite{fischer2011history} as:
\begin{equation}
    n_\text{normal} = \left( \frac{z_{\alpha/2} \cdot \sigma}{d} \right)^2,
\end{equation}
where $z_{\alpha/2}$ is the z-score corresponding to the significance level $\alpha$ under a normal distribution.
However, since only a fraction $p$ of the total samples yield nonzero activations, the expected number of usable samples is $p\cdot n_\text{normal}$. Therefore, the sample size for estimating the mean activations of sparse features is adjusted as:
\begin{equation}
    n_\text{sparse} = \frac{1}{p} \left( \frac{z_{\alpha/2}\cdot \sigma}{d} \right)^2.
\end{equation}
This result highlights the influence of feature sparsity: as $p$ decreases, the total number of required samples grows proportionally as $1/p$.
In practice, if a task-specific feature occurs with a probability $p=0.01$ in the documents (which is already an ideal situation), estimating the mean activation within a margin $d=0.1\times \sigma$ at a 95\% confidence level would require approximately: 
\begin{equation}
    n_\text{sparse} = \frac{1}{0.01} \left( \frac{1.96\times \sigma}{0.1\times \sigma} \right)^2 \approx 38416.
\end{equation} 
Given that real-world datasets~\cite{wang2018glue} usually consist of thousands of training samples, which is significantly smaller than $n_\text{sparse}\approx 38416$, effectively estimating the average activations $\mathbf{A}$ of the learned sparse feature is infeasible in practice.

\begin{table*}[t]
\small
    \centering
    \caption{Additional qualitative analysis on LLM-identified intended/unintended features for the Reward Modeling task, where the classifier predicts the helpfulness of a chatbot's response to the user. We \colorbox{pink}{highlight} phrases in the text spans that are semantically correlated to helpful behaviors, such as ``being polite'', ``considering diverse cultures'', and ``solving math problems''.}
    \label{rm_case_study}
    \vspace{-0.3cm}
    \begin{tabular}{p{11.0cm}|p{4.0cm}|c}
        \toprule
        \toprule
        \centering
        \textbf{Most Activated Text Spans} & \textbf{Summary of Text Spans} & \textbf{Helpful Relevant} \\
         \midrule
        \textbf{Span  1:} \_[BOT]:\_\_ \colorbox{pink}{It could}; \textbf{Span 2:} \_\_ \colorbox{pink}{Is it possible} that your symptoms could be related; \textbf{Span 4:} \_\_[BOT]:\_\_ \colorbox{pink}{Perhaps} there’s; \textbf{Span 4:} BOT]:\_\_ Good question! \colorbox{pink}{It might be}; \textbf{Span 5:} [BOT]:\_\_ Do you think \colorbox{pink}{it might}; \textbf{Span 6:} s a possibility of damage, but it \colorbox{pink}{could also}; \textbf{Span 7:} BOT]:\_\_ If you are, \colorbox{pink}{you might}; \textbf{Span 8:} n\_\_[BOT]:\_\_ \colorbox{pink}{Maybe} you’; \textbf{Span 9:} n\_\_[BOT]:\_\_ It \colorbox{pink}{could be}; \textbf{Span 10:} \_\_[BOT]:\_\_ \colorbox{pink}{Could it} & Expressing a possibility or uncertainty. & \multirow{5}{*}{Yes}\\ \hline
        \textbf{Span 1:}  a \colorbox{pink}{Japanese} mayonnaise called “Nozaw; \textbf{Span 2:}  is popularly served with \colorbox{pink}{Katogo} (beef; \textbf{Span 3:} ame, \colorbox{pink}{Spyro}: Year of the Dragon; \textbf{Span 4:}  upgrade pack called \colorbox{pink}{Diablo II}: Lord of Dest; \textbf{Span 5:}  two hours after giving the infant iron-containing; \textbf{Span 6:}  also listen to \colorbox{pink}{George Harrison’s} "My Sweet; \textbf{Span 7:} ina is \colorbox{pink}{Tina Turner’s} autobiography; \textbf{Span 8:}  model name, for example the 901; \textbf{Span 9:} I suggest looking for pieces called \colorbox{pink}{Goldberg Vari}; \textbf{Span 10:}  a message like “The bear ate the gummy & The provided text Spans contain references to diverse cultural items (Japanese mayonnaise), food (Katogo), and video games (Spyro: Year of the Dragon, Diablo II: Lord of Destruction). &  \multirow{7}{*}{Yes}\\ \hline
        \textbf{Span 1:}  day will be \colorbox{pink}{24 hours}, 1; \textbf{Span 2:}  \colorbox{pink}{20 years}, you will have \$1; \textbf{Span 3:}  \colorbox{pink}{37 degrees Fahrenheit = 3}; \textbf{Span 4:}  Sure. \colorbox{pink}{3.5 rounded up} is ; \textbf{Span 5:}  \colorbox{pink}{37 degrees Fahrenheit = 3} & The feature captures numerical values and conversions. & \multirow{2}{*}{Yes}\\ \hline
        \textbf{Span 1:}  Was considering; \textbf{Span 2:}  I'm thinking of getting; \textbf{Span 3:}  I was thinking about getting; \textbf{Span 4:}  I'm thinking about getting; \textbf{Span 5:}  What should I look for; \textbf{Span 6:}  Why are there; \textbf{Span 7:}  What things should I look for; \textbf{Span 8:}  Should I get; \textbf{Span 9:}  I'm trying to decide whether; \textbf{Span 10:}  How do I decide between getting & The feature captures user general queries. & \multirow{4}{*}{No}\\ 
        \bottomrule
        \bottomrule
    \end{tabular}
    \label{tab:qualitative}
    \vspace{-0.4cm}
\end{table*}

\section{Details of Training Sparse Autoencoders}
\label{appd:training_sae}

\subsection{Pre-training Sparse Autoencoders.} 
The pre-training process and hyperparameter configuration adhere to established practices outlined in prior research~\citep{brickentowards,gao2024scaling,lieberum2024gemma}. For pre-training, we curated data from diverse instruction-tuning datasets, including ShareGPT~\citep{shareGPT}, UltraChat~\citep{ding2023enhancing} (randomly sampling 400,000 instances), HH-RLHF~\citep{bai2022training}, WebGLM-QA~\citep{liu2023webglm}, Evol-Instruct~\citep{xu2023wizardlm}, and HelpSteer2~\citep{wang2024helpsteer2}, while removing duplicate prompts to produce a total of approximately 711,000 unique queries spanning various topics and intents. This dataset was split into training (90\%) and validation (10\%) sets, yielding around 113 million tokens for training and 12 million tokens for validation, with an average query length of approximately 178 tokens. We initialize the sparse autoencoder with $2^{16}$ feature vectors using Kaiming initialization~\citep{he2015delving}, selecting this number based on the scaling law, i.e., $C = \mathcal{O}(Z^\gamma)$, between the feature count $C$ and the training tokens $Z$ observed in~\cite{gao2024scaling}, with $\gamma \approx 0.60$ for GPT2-small, $\gamma \approx 0.65$ for GPT-4, and $\gamma \approx 0.5978$ in our analysis. We also set Top-$K=20$ feature vectors used for each inference following previous work~\cite{gao2024scaling}. 
We train the sparse autoencoder over five epochs on the pre-training dataset with a batch size of 512 prompts.

\subsection{Fine-tuning Sparse Autoencoders.} 
We fine-tune the pre-trained sparse autoencoder on the training dataset of each task. 
At the beginning of each fine-tuning epoch, we use all feature vectors to go through the entire dataset and monitor those feature vectors that have not been activated as dead neurons. 
The Top-$K=20$ dead feature vectors are asked to reconstruct the residuals of text representations with $\alpha=0.1$. 
Fine-tuning is performed using the AdamW~\cite{loshchilov2017fixing} optimizer with $\epsilon=6.25 \times 10^{-10}$, $\beta_1=0.9$, and $\beta_2=0.999$. 
For the Toxic Chat Detection and Reward Modeling tasks, fine-tuning is conducted for a total of 5 epochs with a batch size of 512 and a learning rate of $5 \times 10^{-5}$. 
In contrast, for the Disease Diagnosis task, fine-tuning runs for 40 epochs with a batch size of 8 and a learning rate of $3 \times 10^{-6}$, as the training dataset is significantly smaller than those of the other two tasks.

\section{Details of Identifying Unintended Features}
\label{appd:interpret_sae}
\subsection{Interpreting Learned Features}
Building on prior work in LLM-as-a-judge~\citep{bills2023language,chaudhary2024evaluating,gao2024scaling,lieberum2024gemma}, we interpret the learned feature vectors from fine-tuned sparse autoencoders by identifying the Top-10 text spans that most activate each feature, with each span limited to a maximum of 32 tokens. To summarize the underlying patterns of these activations, we employ gpt-4o-mini-2024-07-18 as our machine annotator, with a temperature of 0 for deterministic decoding. Each generated response is capped at 1024 tokens. 
To enhance the reliability of this process, we design a structured prompting framework that incorporates a role-playing strategy and in-context examples. Following previous work~\citep{bills2023language}, our machine annotator has the option to respond with ``Cannot Tell'' if it detects no meaningful patterns among the activated text spans. Additionally, to mitigate hallucinations, we prompt the LLM in a separate thread to verify whether its previously generated summary accurately reflects the underlying data. 

\subsection{Identifying Unintended Features}
For all three tasks, we define unintended features as those that lack a clear semantic correlation with the task, as determined by human-crafted evaluation rubrics. Specifically, we reference prior work to establish these rubrics: toxic detection follows~\citep{dubey2024llama3herdmodels}, reward modeling is based on~\citep{ouyang2022training}, and disease diagnosis adheres to guidelines from~\citep{xu2019end}. 
To assess feature-task relevance, we adopt the annotation framework from~\citep{lieberum2024gemma}, where the machine annotator classifies correlations into four levels: ``Yes'', ``Probably'', ``Maybe'', and ``No''. Features are considered unintended if they are are assigned a task relevance level lower than ``Probably''.
\section{Extended Case Study on Reward Modeling}
\label{appd:case_study}
Table~\ref{tab:qualitative} provides an extended qualitative analysis of LLM-identified intended and unintended features for the Reward Modeling task. In this task, the classifier predicts the helpfulness of a chatbot's responses to users. Unintended features are defined as those that are not semantically aligned with the goal of classifying helpfulness. 
A particularly noteworthy example is the second feature, where the raw explanations highlight a diverse range of cultural references.
Conversely, other features such as the fourth example (focused on repetitive phrases like “over the”) are identified as unintended, as they lack meaningful alignment with the task. 

\end{document}